\documentclass[conference]{IEEEtran}
\IEEEoverridecommandlockouts

\usepackage{cite}
\usepackage{amsmath,amssymb,amsfonts}
\usepackage{algorithmic}
\usepackage{graphicx}
\usepackage{textcomp}
\usepackage{xcolor}
\usepackage{tikz}
\usepackage{float}
\usepackage{tcolorbox}

\usepackage{pifont}
\newcommand{\cmark}{\ding{51}}%
\newcommand{\xmark}{\ding{55}}%
\usepackage[table]{xcolor}
\usepackage{tabularx}
\usepackage{enumitem}
\definecolor{green}{HTML}{66FF66}
\definecolor{myGreen}{HTML}{009900}
\def\BibTeX{{\rm B\kern-.05em{\sc i\kern-.025em b}\kern-.08em
    T\kern-.1667em\lower.7ex\hbox{E}\kern-.125emX}}
\begin{document}

\title{Uncertainty Awareness and Trust in Explainable AI - On Trust Calibration using Local and Global Explanations\\

\thanks{Research Center Trustworthy Data Science and Security, TU Dortmund}
}

\author{\IEEEauthorblockN{Carina Newen
\IEEEauthorblockA{\textit{Research Center Trustworthy Data Science and Security} \\
TU Dortmund University\\
carina.newen@cs.tu-dortmund.de\\
0000-0001-8721-6856}}
\and
\IEEEauthorblockN{Daniel Bodemer}
\IEEEauthorblockA{\textit{University Duisburg-Essen} \\
0000-0003-2515-683X}
\and
\IEEEauthorblockN{Sonja Glantz}
\IEEEauthorblockA{\textit{University Duisburg-Essen}}
\and
\IEEEauthorblockN{Emmanuel Müller}
\IEEEauthorblockA{\textit{Research Center Trustworthy} \\
\textit{Data Science and Security}\\
TU Dortmund University\\
0000-0002-5409-6875}
\and
\IEEEauthorblockN{Magdalena Wischnewski}
\IEEEauthorblockA{\textit{Research Center Trustworthy} \\
\textit{Data Science and Security}
0000-0001-6377-0940}
\and
\IEEEauthorblockN{Lenka Schnaubert}
\IEEEauthorblockA{\textit{University of Nottingham} \\
0000-0002-6863-8110}
}

\maketitle

\begin{abstract}
Explainable AI has become a common term in the literature, scrutinized by computer scientists and statisticians and highlighted by psychological or philosophical researchers. One major effort many researchers tackle is constructing general guidelines for XAI schemes, which we derived from our study. While some areas of XAI are well studied, we focus on uncertainty explanations and consider global explanations, which are often left out. We chose an algorithm that covers various concepts simultaneously, such as uncertainty, robustness, and global XAI, and tested its ability to calibrate trust. We then checked whether an algorithm that aims to provide more of an intuitive visual understanding, despite being complicated to understand, can provide higher user satisfaction and human interpretability. 
\end{abstract}


\section{Introduction}
In recent years, the efforts regarding machine-learning-assisted research have doubled and tripled, which has led to artificial intelligence applications being introduced into high-risk domains such as credit scoring \cite{dastile2020statistical}, medical diagnosis \cite{kononenko2001machine} or threat detection \cite{mayhew2015use}. It has become a primary directive to ensure the trustworthiness of current and future technologies (e.g., European Commission 2019, IBM). A standard tool for establishing trust has been explainable artificial intelligence (XAI). While we see efforts from the psychological side to provide general guidelines for XAI research \cite{dieber2022novel} or attempts to measure the usability of an algorithm on more minor scales for some proposed algorithms \cite{ribeiro2016should}, it has yet to become accepted that newly established explainability methods are evaluated and improved regarding the effectiveness of their trust calibration ability. It is common that from the computer science community, human interpretability is simply claimed or evaluated at really small scales with very few test subjects \cite{ribeiro2016should}. While this is often the only thing that can be done without proposing a large-scale human study every time for each algorithm, we recognize that detailed studies are necessary, at least for representatives. As we show here later, with our study, we were able to identify the essentials of a global XAI algorithm in the uncertainty domain without producing cognitive overload for the users. The concept we found can be extended to other algorithms potentially making the same mistakes, but we would not have been able to identify it without this in-depth study with open questions.

While analyzing every single algorithm on its own also comes with enormous effort and perhaps little insight, we show that there is a knowledge gain given if we do not just analyze the most common algorithms but other representatives that are state-of-the-art in their own niche. For specific purposes, we still need new algorithms. To give an example: While LIME \cite{ribeiro2016should}, Integrated gradients \cite{sundararajan2017axiomatic} and its countless extensions cover locally faithful explanations, they have informatory value on how well an algorithm performs overall on an entire dataset. However, it might be helpful to evaluate a machine learning model on a global scale rather than only on local examples. The same can be said about other commonly used explanation algorithms: Shapley values \cite{lundberg2017unified} are often not considered human interpretable (at least regarding non-experts in the field) \cite{kumar2020problems}. Analyzing only the most commonly used XAI methods, we might not learn anything new regarding methods with different goals, inputs, and details compared to the most popular methods, which means we will not spot important usability concerns. Other attempts to include more reliability/ robustness notions are, for example, a focus on the stability of the explanation, however again, without focusing on the robustness notion \cite{slack2021reliable}. In Table \ref{tab:comparison}, we give an overview of some of the algorithms that tackle similar goals as our chosen algorithm and why we chose this particular one. 
\begin{table}[ht]
\centering
\caption{Overview of algorithm choices and their related concepts }
\scalebox{0.6}{%
\begin{tabular}[t]{lccccc}
\hline
 & global & model-agnostic & Uncertainties & Robustness & usable for non-experts\\
\hline
LIME  \cite{ribeiro2016should} & \xmark & \cmark & \xmark & \xmark & \cmark \\
SHAP \cite{lundberg2017unified} & \cmark & \cmark & \xmark & \xmark & \xmark\\
Integrated gradients \cite{sundararajan2017axiomatic} &  \xmark & \xmark & \xmark & \xmark & \cmark \\
Uncertainty DeepSHAP \cite{marx2023but} & \xmark & \cmark & \cmark  & \xmark &  \cmark \\
CLUE \cite{clue} & \xmark & \xmark & \cmark & \xmark & \cmark \\
Bayesian Local Explanations \cite{slack2021reliable} & \xmark &  \cmark &  \cmark & \xmark & \xmark \\
Unsupervised DeepView \cite{newen2022unsupervised} & \cmark & \cmark & \cmark & \cmark & \cmark\\

\hline
\end{tabular} }
\label{tab:comparison}

\end{table}%
\\
While this might only be feasible for some upcoming approaches, we chose a less studied area of XAI regarding the uncertainty explanation domain. Rather than going by popularity, we chose a representative from that area that covers a variety of properties present in the XAI taxonomy. We focused on uncertainties because it has been shown that explanations that are not uncertainty-aware can often be inconsistent \cite{marx2023but}. The common approach tackled by most XAI researchers includes works on how to extend state-of-the-art methods to incorporate explanations \cite{slack2021reliable} or why XAI and Uncertainties combined help to calibrate trust \cite{seuss2021bridging}. We consider global explanations to be explanations that do not just cover single instances and a model's performance on those, but overall estimate the explainability on an entire dataset \cite{setzu2021glocalx}. It is important to get more than just local snapshots of a model's trustworthiness. Model-agnostic states that the explanation method does not require prior knowledge of the models' internals \cite{setzu2021glocalx}, which was already recognized as a desirable metric in the very first works on explainable AI \cite{ribeiro2016should}, and has since been established \cite{kumar2020problems}. The robustness that we estimate with this explanation method denotes the likelihood of adversarial attacks \cite{newen2022unsupervised}. Adversarial attacks describe instances that are altered in an inconspicuous manner but then misclassfied \cite{goodfellow2015explainingharnessingadversarialexamples}.

We define trust calibration as the balance between human trust and the actual trustworthiness of the application \cite{mcdermott2019practical}. We focus on Unsupervised DeepView \cite{newen2022unsupervised}, the first unsupervised explainable AI algorithm for high dimensional data that portrays uncertainty not just by providing conventional uncertainty estimation but also by including a robustness notion. The appeal of this method is that it should help laypersons and experts directly compare the quality of one model with the quality of another model for a set task by giving a visual comparison of certainty.

To improve the general understanding of XAI method improvements, we propose that it is not enough to evaluate a very general setup of all common XAI methods. Suppose that we want to evaluate new methods that fulfill niche purposes that are not covered by the most popular methods. In that case, we have to evaluate those methods in detail and then derive general recommendations from researching the usability and trust calibration abilities of specific newly proposed methods. Hence, in this study, we evaluate the explainable AI algorithm Unsupervised DeepView \cite{newen2022unsupervised} regarding trust calibration and explanation satisfaction using the Explanation Satisfaction Scale \cite{hoffman2018metrics} as an additional evaluation scheme for the quality of the explainability method. We check our trust evaluation using some aspects of the Trust in Automation scale \cite{pohler2016itemanalyse}. In conclusion, we derive some general guidelines for XAI algorithms from the open questions we designed.
\section{Related Work}
\paragraph{Explainable AI} While AI methods are already widely employed, many individuals find it hard to trust an instance without transparency, where its internal workings are a black box. Using explainable AI for trust calibration is generally a known tool, \cite{shin2021effects}. Still, we can gain new insights into the general design of explainable AI (XAI) systems by focusing on trust calibration in specific settings. It allows us to control the environment and focus on details that can lead to bad algorithm designs in future settings. Trust has increasingly become an important concept regarding human-AI (artificial intelligence) interaction \cite{glikson2020human}. There have been many attempts in the literature to formalize the prerequisites and goals of the cognitive mechanism of trust \cite{jacovi2021formalizing}. Literature often distinguishes between cognitive trust and trust based on affect \cite{mcallister1995affect, gillath2021attachment}. Other approaches, rather than just formalizing it, which attempt to understand trust, have dealt with calibration and how to influence trust once established \cite{siau2018building, salem2015would}. Furthermore, attempts to map trust onto guidelines for trustworthy machine-learning technologies focus on principles such as fairness, explainability, audibility, and safety \cite{toreini2020relationship}. We will be focusing on explainability and uncertainty, with uncertainty including safety aspects, but not strictly covering the same aspects. Other dimensions of Trust in AI include tangibility \cite{bainbridge2011benefits, lee2006can } and transparency \cite{dzindolet2003role}, often attempted using explainability methods. The most common explainable AI methods include visualization techniques such as LIME \cite{ribeiro2016should}, Shapely values \cite{lundberg2017unified}, Anchors \cite{ribeiro2018anchors} and LIME-based extensions such as EXPLAIN-IT \cite{morichetta2019explain}, which is used in unsupervised learning contexts. The background for explainable AI solutions is extensive: While LIME provides intuitive visualizations and is based on local machine learning model estimation, solutions like shapely values find their origin in game theory and compute the average marginal contribution of a feature value across all possible coalitions.  
In addition to these local explainability attempts, methods like DeepView \cite{schulz2019deepview} focus on visualizations of global decision boundaries of deep neural networks. This was later extended to Unsupervised DeepView \cite{newen2022unsupervised}, an algorithm that uses local intrinsic dimensionality to visualize unsupervised decision boundaries. Other methods employing global rather than local explanations include feature importance visualizations \cite{casalicchio2019visualizing} or do not apply to deep neural networks \cite{schulz2019deepview}. From the technical side, LIME \cite{ribeiro2016should} has already identified the essential features for an explainability algorithm: model-agnosticism, local faithfulness, and human interpretability. The only way to evaluate human interpretability is to test them and their trust calibration effectiveness using machine learners in user studies. However, this is rarely done for new models, and even less often, we derive practical and not just theoretical guidelines for the design of future XAI algorithms. In this study, we will be doing both.  

\paragraph{User-center XAI (Explainable AI)}
In the past, the efforts to connect explainable AI with user-centred research include research regarding the evaluation of LIME itself \cite{dieber2022novel}, where the framework is explicitly derived for this baseline explainability method and more abstract evaluation schemes such as general usability principles for XAI \cite{kirsch2017explain}. Furthermore, Lim and Dey et al. \cite{lim2009and} show that explaining why the system behaves in a particular manner leads to a better understanding of the system than showing why it does not. In addition, other researchers have highlighted how user-centered approaches shape or should shape XAI \cite{mohseni2021multidisciplinary}. Some user-centered AI evaluations focus on targeting specific groups of users and present solutions \cite{ribera2019can}. Even other work focuses on the importance of XAI and its interrelation to terms like interpretability, transparency, explicitness, and faithfulness \cite{rosenfeld2019explainability}. A fair amount of systematic reviews exist on XAI and the applicability of certain XAI systems to specific scenarios \cite{anjomshoae2019explainable, arrieta2020explainable}. Liao et al. \cite{liao2020questioning} also provided informed design practices, which we also applied in developing our scenario. Therefore, we extend their recommendations by testing how well trust calibration works for practical AI algorithms when following those guidelines. What makes this user study unique is the focus on a practical uncertainty explanation method that provides guidelines and evaluations for a combination of local instances and global explanations. A combination of the two might be a valuable recommendation for future algorithms, as it allows intuitive evaluation of single instances, and provides an overview of how they will perform on an entire dataset \cite{lundberg2020local}. Furthermore, it gives a fair evaluation of user satisfaction and the trustworthiness of an existing explainability method.

\section{Experiment}
We tested three hypotheses: 
\begin{itemize}
    \item Unsupervised DeepView supports trust calibration.
    \item Unsupervised DeepView performs better than numerical methods regarding satisfaction in the explanation.
    \item Unsupervised DeepView is more trustworthy than a numerical method.
\end{itemize}

Ideally, trust calibration should be the ability to let the human perception be identical or at least near-identical to the trustworthiness of the learner itself \cite{mcdermott2019practical}. However, verifying this alignment is practically unattainable. Instead, we investigated if objectively good learners are statistically trusted more than bad ones. This served as a foundational test; failure would invalidate subsequent experiments. For the second hypothesis, we wanted to show by example that even complex explanations might lead to higher satisfaction if the user feels he understood the learner's capabilities more. We also checked the trust calibration ability of a complicated explanation versus a numerical method, which would be the simplest possible alternative. We designed some open questions to see whether we could derive general guidelines for other XAI algorithms and were able to do so. To this aim, participants were presented with a short scenario about classification algorithms, and their task was to evaluate them on several dimensions. The crucial manipulation was whether only a numeric explanation was provided, or additionally the Unsupervised DeepView explanation. To answer the research questions and test the hypotheses, an online study was conducted via Lime Survey. A one-factorial between-subject design with two levels on the factor was used for the study. The two factors were the numeric or visual conditions for the two different learners. Numeric denotes that participants were exposed to training and validation accuracies, and the visual group received both a numerical explanation and Unsupervised Deepview as a visual explanation. We performed one baseline check, and the two scales, namely the explanation satisfaction scale and trust in automation scale, were evaluated by averaging all results of the separate items, as suggested in their design.
\subsection{Participants}
Subjects were recruited through the online crowdsourcing platform "Prolific". This allowed subjects to receive monetary compensation for their participation. Participants received 6 euros for 30 mins average participation time. They had to be at least 18 years old and fluent in German to participate in the study. Initially, they were informed about their rights and obligations regarding the survey. The study was approved by the ethical commission of the TU Dortmund university.
A total of 265 data sets were generated. After cleaning, 196 records remained. We had previously aimed for ~190 participants in order to satisfy statistical significance. We estimated this amount of people using power analysis and settling on $\alpha=0.05$ as well as the statistical power $=0.8$. This gives us enough statistical significance with our sample size, and luckily our measured Cohen's d effect sizes are high enough when a statistical significance is found. Primarily, English and incomplete answers were excluded. In the final sample, there were 96 subjects in the control group (numeric) and 100 in the experimental group (visual). There were 100 male participants, 93 female, and three diverse participants. The participants were 18 to 73 years old, and the mean age was M= 31.03 (SD = 10.5). The distribution of the genders per group was 55\% men vs. 44\% women vs. 1\% diverse for the group with visual explanations and 47\% men vs. 51\% women vs. 2\% diverse. 
For the task, participants received different information according to the experimental group. The numeric group had to base their decision on accuracy percentages, often used to evaluate artificial intelligence when you have no specialized knowledge in the field or additional explanations \cite{dong2015knowledge}. The visual group received the percentages and the visual explanations that provided global and local explainability. The visualization method used for our study is Unsupervised DeepView \cite{newen2022unsupervised}.  
\subsection{Design of the Scenarios}
\begin{figure}[t]
\centering
\includegraphics[width=\columnwidth]{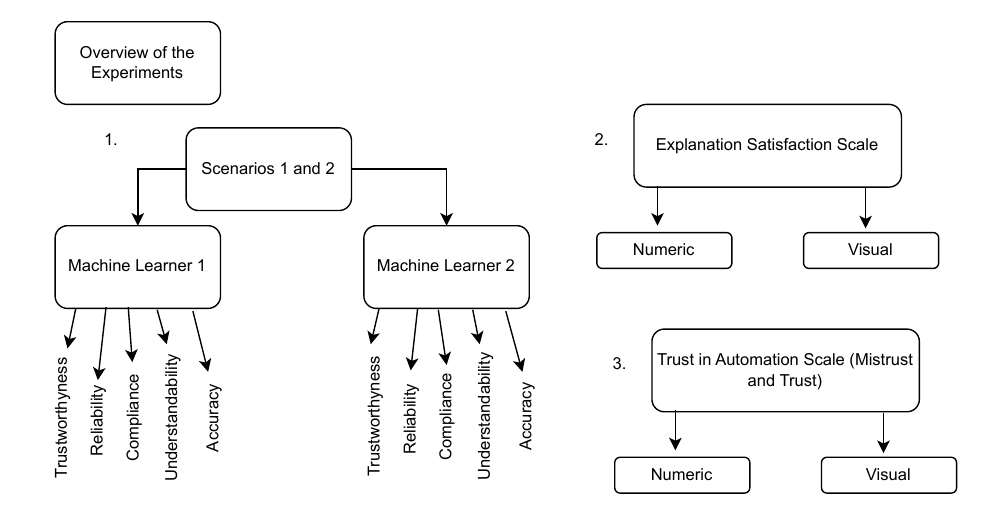} 
\caption{Overview of the key components of this paper.}
\label{fig:Overview}
\end{figure}

Liao et al. \cite{liao2020questioning} provided informed design practices, which we applied in developing our scenario. The explanations given to the study participants were evidence-based: It was possible to see instances of data points from the MNIST number dataset \cite{deng2012mnist} that were deemed uncertain. The participants could derive from those instances whether they looked hard to classify or whether the numbers were cleanly written and should be recognized instantly. Also, we made sure that users made an informed decision: While it is impossible to explain all concepts behind artificial intelligence in a short paragraph, we made sure that the users understood what the explanation's implications are: Unsupervised DeepView deems an image to be uncertain if either, the model itself is uncertain about its decision, or, the likelihood that the image is manipulation and therefore an attack on the network, is high. This means that we include a notion of robustness in the explanations. After receiving general instructions, all participants were presented with a short scenario. We constructed scenario one, which acted as a baseline. If participants' trust levels were not measurably different for objectively better or worse models, even when using simple numeric methods, measuring trust calibration on more complicated explanations might have been futile. The second scenario tests the trust calibration on the visual explanation of Unsupervised DeepView. For the scenarios, participants were asked to rate different machine learning algorithms in terms of trustworthiness, reliability, competence, accuracy, and understandability. The scenarios were self-developed and written. In addition, we assessed various variables that are related to the quality of AI systems. 
The basis for all decisions was the following scenario:  \\
\\
You are an entrepreneur of a company that wants to switch to systems supported by artificial intelligence for certain tasks. For this, you get some offers from partner companies. They all advertise great services; however, now you must choose the best algorithm. The task of the AI system will be to classify images correctly. In this first example, it simply recognizes handwritten numbers (0-9). They use the MNIST dataset, which consists of handwritten numbers standardized to 28* 28 pixels. It is a dataset commonly used for the evaluation of artificial intelligence. Figure \ref{fig:Overview} shows an overview of the experiments' key components.

\begin{figure*}[t]
\centering
\includegraphics[width=17cm]{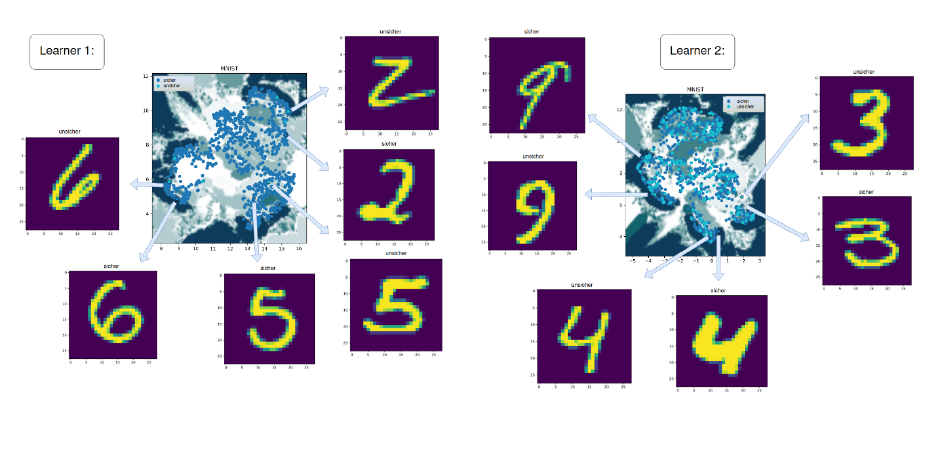} 
\caption{Snapshot of the original study: "unsicher" means insecure and "sicher" translates to "certain". On the left-hand side, we see machine learner one, the right-hand side shows machine learner two. }
\label{fig:snap}
\end{figure*}

\paragraph{\textbf{Scenario 1:}} For the first scenario, we introduced the metrics test and validation accuracy, which are common machine learning metrics \cite{dong2015knowledge}. The first scenario was visible to both the numeric and visual groups. The task for the first scenario was the following: The test subjects were explained the difference between test and validation accuracy and then were asked to distinguish between two learners in the form of a text scenario: Learner 1 had 99\% test accuracy and 80\% validation accuracy, whereas learner 2 had 95\% test accuracy and 90\% validation accuracy. The participants were asked to rate both machine learners separately regarding accuracy, reliability, competence, understandability, and trustworthiness on a 6-point equidistant response format ranging from completely agree to completely disagree. If trust calibration succeeds with this metric, Learner 2 should be trusted more than Learner 1 because very high test accuracy can mean the data was learned by heart and not generalized. 
\paragraph{\textbf{Scenario 2:}} Scenario two was only presented to the visual group: The task was again to rank two learning algorithms for their trustworthiness, reliability, competence, understandability, and accuracy on a 6-point equidistant response format ranging from 1 = completely agree to 6= completely disagree. This time, we provided a global overview of a whole dataset using visual explanations generated by the Unsupervised DeepView algorithm \cite{newen2022unsupervised} and single instances classified as certain or uncertain. The study participants were given the following information: The explanation method shows a global picture of how many data points were deemed as uncertain and on the side examples of those instances. Uncertain means that the machine learner is not robust for this image and is easy to attack or that the learner is unsure about its prediction. Every data point stands for a picture from the data set. The background shows rough estimations of unseen data: White areas mean no data to make a conclusive prediction, dark blue means that Unsupervised DeepView estimates it to be certain, and light blue means we roughly estimate it to be uncertain. The darker the picture is, the more certain the machine learner is overall. Participants now had the choice to either estimate the trustworthiness of learners from the global images or try to make sense of whether the example images labeled as uncertain looked hard to classify. The same could be said about whether easily distinguishable numbers were also labelled as certain. This information, however, was not given in the study; we just explained how the method works and then presented the two machine learners. Both learners with different explanations were given the same overall test- and validation accuracy. The difference was that one learner labelled instances that looked badly written as uncertain, therefore providing a "trustworthy" estimate of uncertain and certain, and the other did the opposite. As can be seen in Figure \ref{fig:snap}, Unsupervised DeepView consists of one global image of all data points and several example images labeled as uncertain or certain. Machine learner one had much fewer global predicted uncertainties as compared to machine learner two. This is because the algorithm on the right was trained without regularization, a machine-learning technique known to have positive effects on machine-learning applications, such as better generalization \cite{sjoberg1995overtraining}. Furthermore, the example instances and the global visualization were real instances generated by Unsupervised Deepview \cite{newen2022unsupervised} to represent the actual behavior of the algorithm as closely as possible while remaining in a controlled setting.   
\begin{figure}[ht]
\label{fig:exsat}

    \centering
    \begin{tcolorbox}[colframe=black!70, colback=white, sharp corners=south]
    \begin{tabularx}{0.95\textwidth}[t]{XXXX}
        \arrayrulecolor{green}\hline
        \textbf{\textcolor{myGreen}{Explanation Satisfaction Scale Items}} \\
        \hline
    \end{tabularx}
    
    \vspace{0.5em} 
    
    \begin{minipage}{0.95\textwidth}
        \begin{itemize}
            \item[1.] Through the explanation, I know how the machine learner works.
            \item[2.] This explanation of how the machine learner works is satisfying.
            \item[3.] This explanation of how the machine learner works has sufficient detail.
            \item[4.] This explanation of how the machine learner works seems complete.
            \item[5.] This explanation of how the machine learner works tells me how to use it.
            \item[6.] This explanation of how the machine learner works is useful to my goals.
            \item[7.] This explanation of how the machine learner works shows me how accurate the learner is. 
            \item[8.] This explanation of how the machine learner works lets me know how trustworthy the learner is.
        \end{itemize}
    \end{minipage}
    \end{tcolorbox}

    \caption{Explanation Satisfaction Scale Items}
    \label{fig:exsat}
\end{figure}
\paragraph{Explanation Satisfaction Scale:}
Following the scenarios, all participants from both groups were asked to answer various questionnaires and provide information about their demographics and prior experience with AI. The numeric group answered those questions only on the numeric explanation, while the visual group was only asked to focus on the visual explanations. We used the Explanation Satisfaction Scale\cite{hoffman2018metrics} to measure satisfaction with the explanations. This scale consists of 8 items to assess the satisfaction of the method to be answered using a 5-point equidistant response format ranging from "completely agree" to "completely disagree". For both scales, we used the point scale suggested in the respective design paper. For the study, the items were translated into German. Reliability was adequate- with a Cronbach's alpha of 0.746. For an overview, the items of the explanation satisfaction scale are given in Figure \ref{fig:exsat}. 
\\
\paragraph{Trust in Automation Scale:}
Trustworthiness was assessed using the German version of the Trust in Automation Scale\cite{pohler2016itemanalyse}. The scale contains 11 items, and the subscales trust (6 items) and distrust (5 items) are to be answered using a 7-point equidistant response format ranging from "completely agree" to "completely disagree." Due to a technical error in the study, only 10 items were assessed. The missing item from the original scale is "The explanation provides security," which was originally part of the trust scale. The items were conceptually adapted to the context of the study. Reliability was adequate for trust (Cronbach's alpha = 0.729) for trust and slightly lower for distrust (Cronbach's alpha = 0,684) for Cronbach's alpha for distrust, and the subscales correlated by –0.6 using Pearson's r correlation coefficient. The items of the Trustworthiness scale used are given in Figure \ref{fig:trust}.

\begin{figure}[ht]
\begin{tcolorbox}[colframe=black!70, colback=white, sharp corners=south]

\begin{center}
\begin{tabularx}{\textwidth}[t]{XXXX}
\arrayrulecolor{green}\hline
\textbf{\textcolor{myGreen}{Trust in Automation Scale: Mistrust Items}} \\
\hline
\end{tabularx}
\end{center}
\begin{itemize}
    \item[1.] This explanation is misleading.
    \item[2.] This explanation acts obscurely.
    \item[3.] I mistrust the decisions of the explanation.
    \item[4.] I have to be careful regarding the use of the explanation.
    \item[5.] The actions of the explanation can have negative consequences.
\end{itemize}
\begin{center}
\begin{tabularx}{\textwidth}[t]{XXXX}
\arrayrulecolor{green}\hline
\textbf{\textcolor{myGreen}{Trust in Automation Scale: Trust Items}} \\
\hline
\end{tabularx}
\end{center}
\begin{itemize}
    \item[1.] The explanation works without mistakes.
    \item[2.] This explanation is reliable.
    \item[3.] This explanation is trustworthy.
    \item[4.] I can trust this explanation.
    \item[5.] I am familiar with the explanation.
\end{itemize}
\caption{The Trust in Automation Scale}
\label{fig:trust}
\end{tcolorbox}
\end{figure}

\subsection{Open-Format Questions}
Finally, we asked open questions for both groups of participants for a deeper understanding of how people relate to AI, what they currently want from AI, and how difficult/easy each concept is to understand, especially for lay people. The questions were: Which aspects of the explanation did you understand best? Which aspects were most obscure? What would you additionally wish for to improve the explanations? Lastly, we asked what part of the explanations influenced their decision regarding trustworthiness the most: The amount of uncertain pixels or whether the labels "certain" or "uncertain" made sense the most for the group that saw the visualizations. For the numeric group, we just asked what they based their decision on most – training or validation accuracy. Demographics, age, and gender were chosen to provide a representative description of the sample group. Prior experience with AI was surveyed to identify any difference in understanding between the experimental groups. 
\section{Results and Interpretation}
\subsection{Evaluation 1 – Numeric explanation: Do percentage estimations lead to useful trust calibration?}
This evaluation was the baseline check for hypothesis 1. We evaluated using a paired t-test (two-tailed) to give us a first rough estimation of trustworthiness. Reliability was good, with Cronbach's alpha between 0.882 and 0.883. On learner two, Cronbach's alpha= 0.839. The overall evaluation of the items as a scale results in Table \ref{tab:eval1}.

\begin{table}[h]
\centering
\begin{tabular}[t]{ccccc}
\hline
t(195)  & p & mean difference & SE difference & effect size\\
\hline
9.73  & $<$ 0.001  & 0.581 & 0.06 & 0.695 \\

\hline
\end{tabular}
\caption{Do percentage estimations lead to useful trust calibration? }
    \label{tab:eval1}

\end{table}

\noindent In this Table, the SE difference denotes the standard error difference, and the effect size was measured using Cohen's d effect size. The mean of machine learner one was M= 2.947 compared to M=2.366 for machine learner two. The standard deviation of machine learner one was 0.8442. Note that a low value indicates more trust, so the second learner was statistically recognized as more trustworthy overall and regarding all separate items, which seems logical given the provided numeric information. Because this evaluation does not consist of an established scale, we provide an overview of all items not just as a scale but also as single items against one another. We will later evaluate trust again using the Trust in Automation scale in order to have more reliable evidence. 
\begin{figure}
    \centering
    \includegraphics[width=0.99\linewidth]{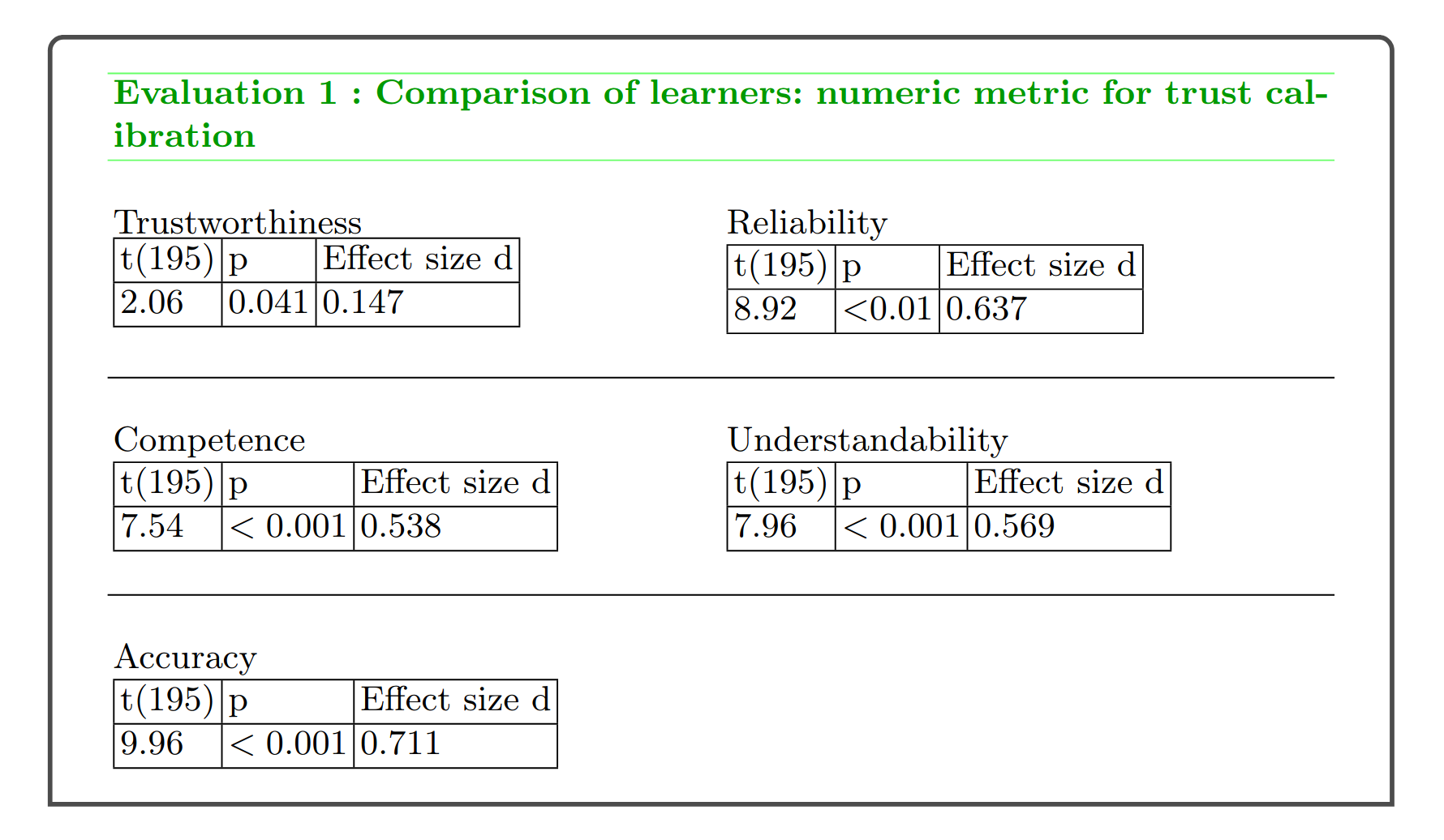}
   \caption{Evaluation of Scenario 1 }
\label{fig:Eval1}
\end{figure}
As we can see here, we detected statistically significant differences between all evaluated factors for the numeric metric. This leads us to assume that trust calibration is at least somewhat feasible, even with numeric measurements. This scenario was not used to evaluate one of the hypotheses but rather provided a baseline on whether trust calibration could be measured using these scenarios. All biases lead to the objectively "better model," with higher validation accuracy yet still good overall test accuracy, to be chosen. In summary, we tested the same key items for a second scenario using Unsupervised DeepView. 

\subsection{Evaluation 2 – Visual explanation: Does Unsupervised DeepView lead to useful trust calibration?}
We evaluated Scenario 2 (see Section Scenario 2) between the two learners across participants of the visual group only (within-subject) using a paired t-test (two-tailed) to give us a first rough estimation of trustworthiness. The reliability analysis of learner one via Cronbach's alpha resulted in 0.918. On learner two, this resulted in 0.895. The overall evaluation of the items as a scale is shown in Table \ref{tab:eval12}.

\begin{table}[H]
\centering
\begin{tabular}[t]{ccccc}
\hline
t(99)   & p & mean difference & SE difference & effect size\\
\hline
-5.17   & $<$ 0.001  & -0.598  & 0.116 & -0.517  \\

\hline
\end{tabular}
\caption{Do percentage estimations lead to useful trust calibration? }
    \label{tab:eval12}

\end{table}
The mean of learner one was 2.912, whereas the mean of the second learner was 3.51. The standard deviation of learner one equaled 1.008. A low value meant more trust, so the second learner was statistically recognized as less trustworthy. This is an expected outcome since we designed learner two to be more uncertain. We conclude that Unsupervised DeepView is capable of some form of trust calibration. The learner who used no regularization was, in all aspects of the scale, significantly different from the much less uncertain model. This means that our hypothesis that Unsupervised DeepView supports trust calibration was confirmed. 
\begin{figure}
    \centering
    \includegraphics[width=0.99\linewidth]{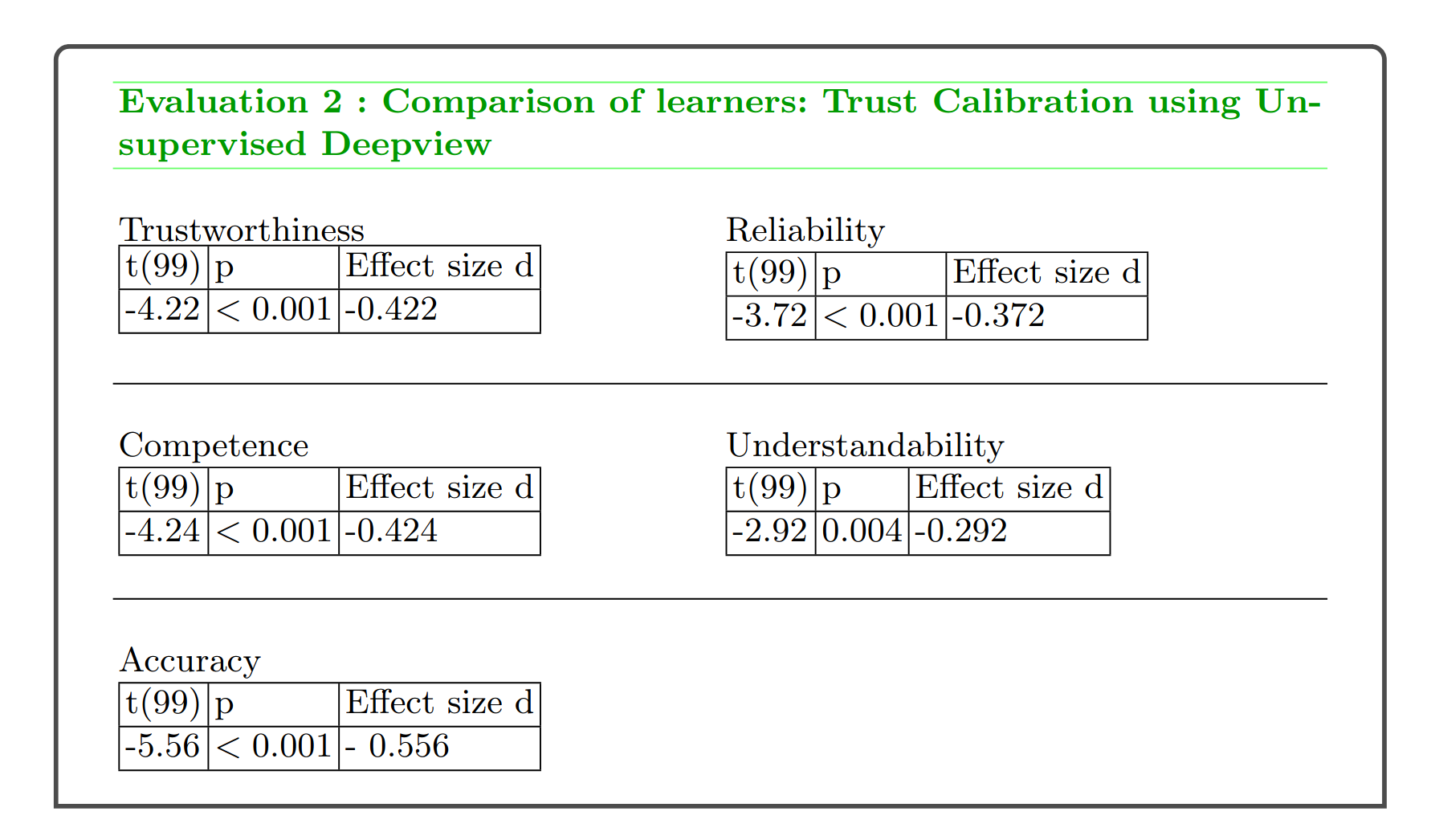}
\caption{Evaluation of Scenario 2}
\label{fig:evalsce2}
\end{figure}

\subsection{Evaluation 3 – Does Unsupervised DeepView lead to more satisfactory explanations than simple numerical values?}
The reliability analysis of the Explanation Satisfaction Scale using Cronbach's alpha was 0.746, which means it was overall good enough for evaluation purposes. We tested the hypothesis regarding explanation satisfaction using an independent sample t-test (assumptions were met: Levene test p=0.306, Shapiro-Wilk p=0.27). Results indicate no difference in satisfaction between Unsupervised DeepView and the numeric visualization and results are shown in Table \ref{tab:eval13}.
\begin{table}[h]
\begin{center}
    \begin{tabular}[t]{ccccc}
\hline
t(194)   & p & mean difference & SE difference & effect size\\
\hline
1.16    &  0.248   & 0.109   & 0.094 & 0.165   \\

\hline
\end{tabular}
\caption{Results of the Explanation Satisfaction Comparison }
    \label{tab:eval13}
    \end{center}

\end{table}
\subsection{Evaluation 4 – Does Unsupervised DeepView lead to more trust in the explanations than simple numerical values? }
Regarding the mistrust items, the reliability analysis of the Risk Factors scale using Cronbach's alpha was 0.684. We performed an independent student t-test for the mistrust scale, resulting in a statistically significant result. 
\begin{table}[h]
\centering
\begin{tabular}[t]{ccccc}
\hline
t(194)   & p & mean difference & SE difference & effect size\\
\hline
-1.82    &  0.070   & -0.26    & 0.143  & -0.26   \\

\hline

\end{tabular}
\caption{Results of the Trust in Automation Scale, Mistrust Items}
    \label{tab:eval14}

\end{table}
The mean for the group with the explanation was 4.05, the mean for the group with only numerical explanation was 4.31, and p =0.070, which means Unsupervised DeepView was mistrusted more. The standard deviation of the experimental group (explanation group) was 1.09. The normality check (Shapiro- Wilk's: p=0.461) and homogeneity check (Levene's: p=0.068) passed. The Reliability Analysis of the trust items was 0.788, despite lacking one item due to technical errors during the study. The normality check failed (Shapiro-Wilk test: W=0.972, p$<$0.001), but the homogeneity test (F=0.0717, p=0.789) did not: Therefore, we used the Mann-Whitney U test, which shows that there was no overall significant difference between Unsupervised DeepView and the numerical explanation regarding the trust factors (Statistic 4766, p=0.933). The t-test statistics for the trust scale are shown in Table \ref{tab:eval15}.
\begin{table}[h]
\centering
\begin{tabular}[t]{ccccc}
\hline
t(194)   & p & mean difference & SE difference & effect size\\
\hline
0.458 & 0.648 & 0.069 & 0.150  & 0.065  \\

\hline
\end{tabular}
\caption{Results of the Trust in Automation Scale, Trust Items}
    \label{tab:eval15}
\end{table}
Interestingly, we can see a different result here for the trust and mistrust scale. They are contrary to our hypothesis three. This could be due to the missing item. However, since reliability is good for the trust scale, other factors may have greater impact. Furthermore, this also supports our speculation that certain questions were answered very negatively regarding Unsupervised DeepView due to the method needing to be simpler. The mistrust scale included items such as "the explanation acts obscurely." Unsurprisingly, a real-world explanation method that takes time to understand is seen as more obscure than a numeric method, which can be explained in two sentences. However, we did not test this and cannot tell from this single evaluation whether this explains all discrepancies. 
\subsection{Evaluation of the Open Question Format }
In our study, we presented three open questions: First, we asked what part of the explanations was easy to understand and which was most obscure. Then, we asked what else users would like to add in order to improve the explanations. Lastly, we asked what part of the explanation was most decisive for them, whether it was the number of insecure pixels or whether the labels of insecure and secure made sense for the single images. Several observations could be drawn from the user's answers: Regarding the explainability method Unsupervised DeepView, the most confusing part of the algorithm seemed to be the estimations in the background. Since the estimations also provide less information than the single instances, a takeaway lesson for other algorithms is that explanations should always stay limited to the minimum amount of information needed to convey the maximum amount of understanding and intelligence. 12\% of all participants exposed to the explanation said they liked the visualizations and found them helpful. However, 5\% remarked that the background color led to considerable confusion. While some participants used the single images to decide whether the uncertain or certain label made sense and based their trustworthiness on that, some only used the global image to base their decision. To quantify what people used as the basis for the trustworthiness of a model, we present the amounts in the following Table 6: 
\begin{figure}[H]
    \centering
    \includegraphics[width=\columnwidth] {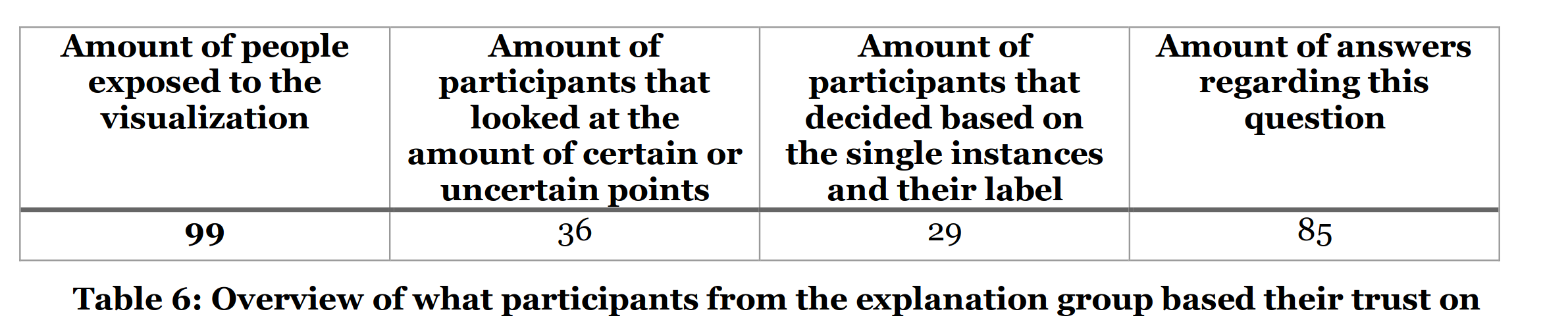}
    \label{fig:tab6}
\end{figure}
This suggests that combining both types of explanations could help ensure clarity. If one explanation is confusing, having both increases the likelihood of understanding while maintaining trust calibration. The visual representation using images was generally well perceived (12 explicitly positive comments). However, both the numerical and visual methods were commented on as needing clarification in some cases. As we know, the explanation satisfaction scale did not show any differences; this reflects that both methods have their weaknesses. Most participants without a background in AI did not know how an artificial intelligence learned at all and, therefore, did not feel comfortable about the explanations or the numerical metrics. One participant explicitly wrote that he does not want or trust artificial intelligence and never will trust it. This observation is not entirely new; it has been noted as trust in familiarity by Tenhundfeld et al.\cite{tenhundfeld2019calibrating}. Furthermore, some participants suggested a video tutorial rather than text to read, which might improve the willingness to use or understand a tool or algorithm. Moreover, one suggestion was that there should be one explanation for the algorithm for an expert group and a less complex one for lay users without background knowledge, which other studies have also found \cite{ribera2019can}. Hence, we rediscovered some suggestions, such as that an optimal explanation might be achieved by explaining a tool in different groups separated by their background knowledge of the field \cite{ribera2019can}. But, we were also able to find out the following proposition: Any XAI algorithm might be improved by leaving out step-by-step the following characteristics: If there is a color scheme, try simplifying it to fewer colors / more clear distinctions and try to find the optimal version that still conveys information but users deem understandable. This was indicated by the background coloring being too complex. 
The visual representations and combination of global and local images, however, seem to have helped people in the decision-making process because, in the setup scenarios, participants trusted the objectively more certain model more (see Evaluation 2), but the choices were not uniquely based on either the local or global explanation, it was roughly even (see Table 6). This indicates that interpretability is aided by a varying explanation for user groups and a variation of local and global explainability.

\section{Limitations and Future Work}
Although we have been careful to systematically vary experimental conditions in the study as well as to use real-world methods, online studies face general constraints. Our sample, though diverse, centred on users of an academic crowd-sourcing platform, encompassing various backgrounds, from novices with scepticism about AI to computer science and machine learning students. Furthermore, the study consisted of (for an online study) a rather long survey time (up to 40 minutes) and required understanding and learning about complex explanation methods. This might have affected the attention spans of some of the participants. To control this factor, we excluded participants who did not pass attention checks from our evaluation. A fundamental difficulty in the experimental design of this study concerns the different amounts of information in the two conditions compared, which may have affected the results. Also, the order of presentation may have affected the participants’ comprehension process. Limitations regarding the survey-based data collection could be considered in future research by using supplementary methods such as eye-tracking in order to evaluate in detail which parts of the explanations the participants focused on most. Also, this work can be used to improve the existing Unsupervised DeepView method and to inform the authors of this paper of the changes that could improve the applicability of their algorithm. Further studies could derive more and more general guidelines for AI explainability algorithms which would help design better algorithms despite the lack of complete user studies for every new niche algorithm.

\section*{Acknowledgment}
This work was funded by the research center trustworthy data science and security (https://rc-trust.ai/). Additionally, this research has been partially funded by the Federal Ministry of Education and Research of Germany and the state of North-Rhine Westphalia as part of the
Lamarr Institute for Machine Learning and Artificial Intelligence.


\begin{thebibliography}{10}
\providecommand{\url}[1]{\texttt{#1}}
\providecommand{\urlprefix}{URL }
\providecommand{\doi}[1]{https://doi.org/#1}

\bibitem{anjomshoae2019explainable}
Anjomshoae, S., Najjar, A., Calvaresi, D., Fr{\"a}mling, K.: Explainable agents and robots: Results from a systematic literature review. In: 18th International Conference on Autonomous Agents and Multiagent Systems (AAMAS 2019), Montreal, Canada, May 13--17, 2019. pp. 1078--1088. International Foundation for Autonomous Agents and Multiagent Systems (2019)

\bibitem{clue}
Antor{\'a}n, J., Bhatt, U., Adel, T., Weller, A., Hern{\'a}ndez-Lobato, J.M.: Getting a clue: A method for explaining uncertainty estimates. arXiv preprint arXiv:2006.06848  (2020)

\bibitem{arrieta2020explainable}
Arrieta, A.B., D{\'\i}az-Rodr{\'\i}guez, N., Del~Ser, J., Bennetot, A., Tabik, S., Barbado, A., Garc{\'\i}a, S., Gil-L{\'o}pez, S., Molina, D., Benjamins, R., et~al.: Explainable artificial intelligence (xai): Concepts, taxonomies, opportunities and challenges toward responsible ai. Information fusion  \textbf{58},  82--115 (2020)

\bibitem{bainbridge2011benefits}
Bainbridge, W.A., Hart, J.W., Kim, E.S., Scassellati, B.: The benefits of interactions with physically present robots over video-displayed agents. International Journal of Social Robotics  \textbf{3},  41--52 (2011)

\bibitem{casalicchio2019visualizing}
Casalicchio, G., Molnar, C., Bischl, B.: Visualizing the feature importance for black box models. In: Machine Learning and Knowledge Discovery in Databases: European Conference, ECML PKDD 2018, Dublin, Ireland, September 10--14, 2018, Proceedings, Part I 18. pp. 655--670. Springer (2019)

\bibitem{dastile2020statistical}
Dastile, X., Celik, T., Potsane, M.: Statistical and machine learning models in credit scoring: A systematic literature survey. Applied Soft Computing  \textbf{91},  106263 (2020)

\bibitem{deng2012mnist}
Deng, L.: The mnist database of handwritten digit images for machine learning research [best of the web]. IEEE signal processing magazine  \textbf{29}(6),  141--142 (2012)

\bibitem{dieber2022novel}
Dieber, J., Kirrane, S.: A novel model usability evaluation framework (muse) for explainable artificial intelligence. Information Fusion  \textbf{81},  143--153 (2022)

\bibitem{dong2015knowledge}
Dong, X.L., Gabrilovich, E., Murphy, K., Dang, V., Horn, W., Lugaresi, C., Sun, S., Zhang, W.: Knowledge-based trust: Estimating the trustworthiness of web sources. arXiv preprint arXiv:1502.03519  (2015)

\bibitem{dzindolet2003role}
Dzindolet, M.T., Peterson, S.A., Pomranky, R.A., Pierce, L.G., Beck, H.P.: The role of trust in automation reliance. International journal of human-computer studies  \textbf{58}(6),  697--718 (2003)

\bibitem{gillath2021attachment}
Gillath, O., Ai, T., Branicky, M.S., Keshmiri, S., Davison, R.B., Spaulding, R.: Attachment and trust in artificial intelligence. Computers in Human Behavior  \textbf{115},  106607 (2021)

\bibitem{glikson2020human}
Glikson, E., Woolley, A.W.: Human trust in artificial intelligence: Review of empirical research. Academy of Management Annals  \textbf{14}(2),  627--660 (2020)

\bibitem{goodfellow2015explainingharnessingadversarialexamples}
Goodfellow, I.J., Shlens, J., Szegedy, C.: Explaining and harnessing adversarial examples (2015), \url{https://arxiv.org/abs/1412.6572}

\bibitem{hoffman2018metrics}
Hoffman, R.R., Mueller, S.T., Klein, G., Litman, J.: Metrics for explainable ai: Challenges and prospects. arXiv preprint arXiv:1812.04608  (2018)

\bibitem{jacovi2021formalizing}
Jacovi, A., Marasovi{\'c}, A., Miller, T., Goldberg, Y.: Formalizing trust in artificial intelligence: Prerequisites, causes and goals of human trust in ai. In: Proceedings of the 2021 ACM conference on fairness, accountability, and transparency. pp. 624--635 (2021)

\bibitem{kirsch2017explain}
Kirsch, A.: Explain to whom? putting the user in the center of explainable ai. In: Proceedings of the First International Workshop on Comprehensibility and Explanation in AI and ML 2017 co-located with 16th International Conference of the Italian Association for Artificial Intelligence (AI* IA 2017) (2017)

\bibitem{kononenko2001machine}
Kononenko, I.: Machine learning for medical diagnosis: history, state of the art and perspective. Artificial Intelligence in medicine  \textbf{23}(1),  89--109 (2001)

\bibitem{kumar2020problems}
Kumar, I.E., Venkatasubramanian, S., Scheidegger, C., Friedler, S.: Problems with shapley-value-based explanations as feature importance measures. In: International conference on machine learning. pp. 5491--5500. PMLR (2020)

\bibitem{lee2006can}
Lee, K.M., Peng, W., Jin, S.A., Yan, C.: Can robots manifest personality?: An empirical test of personality recognition, social responses, and social presence in human--robot interaction. Journal of communication  \textbf{56}(4),  754--772 (2006)

\bibitem{liao2020questioning}
Liao, Q.V., Gruen, D., Miller, S.: Questioning the ai: informing design practices for explainable ai user experiences. In: Proceedings of the 2020 CHI conference on human factors in computing systems. pp. 1--15 (2020)

\bibitem{lim2009and}
Lim, B.Y., Dey, A.K., Avrahami, D.: Why and why not explanations improve the intelligibility of context-aware intelligent systems. In: Proceedings of the SIGCHI conference on human factors in computing systems. pp. 2119--2128 (2009)

\bibitem{lundberg2020local}
Lundberg, S.M., Erion, G., Chen, H., DeGrave, A., Prutkin, J.M., Nair, B., Katz, R., Himmelfarb, J., Bansal, N., Lee, S.I.: From local explanations to global understanding with explainable ai for trees. Nature machine intelligence  \textbf{2}(1),  56--67 (2020)

\bibitem{lundberg2017unified}
Lundberg, S.M., Lee, S.I.: A unified approach to interpreting model predictions. Advances in neural information processing systems  \textbf{30} (2017)

\bibitem{marx2023but}
Marx, C., Park, Y., Hasson, H., Wang, Y., Ermon, S., Huan, L.: But are you sure? an uncertainty-aware perspective on explainable ai. In: International Conference on Artificial Intelligence and Statistics. pp. 7375--7391. PMLR (2023)

\bibitem{mayhew2015use}
Mayhew, M., Atighetchi, M., Adler, A., Greenstadt, R.: Use of machine learning in big data analytics for insider threat detection. In: MILCOM 2015-2015 IEEE Military Communications Conference. pp. 915--922. IEEE (2015)

\bibitem{mcallister1995affect}
McAllister, D.J.: Affect-and cognition-based trust as foundations for interpersonal cooperation in organizations. Academy of management journal  \textbf{38}(1),  24--59 (1995)

\bibitem{mcdermott2019practical}
McDermott, P.L., Brink, R.N.t.: Practical guidance for evaluating calibrated trust. In: Proceedings of the Human Factors and Ergonomics Society Annual Meeting. vol.~63, pp. 362--366. SAGE Publications Sage CA: Los Angeles, CA (2019)

\bibitem{mohseni2021multidisciplinary}
Mohseni, S., Zarei, N., Ragan, E.D.: A multidisciplinary survey and framework for design and evaluation of explainable ai systems. ACM Transactions on Interactive Intelligent Systems (TiiS)  \textbf{11}(3-4),  1--45 (2021)

\bibitem{morichetta2019explain}
Morichetta, A., Casas, P., Mellia, M.: Explain-it: Towards explainable ai for unsupervised network traffic analysis. In: Proceedings of the 3rd ACM CoNEXT Workshop on Big DAta, Machine Learning and Artificial Intelligence for Data Communication Networks. pp. 22--28 (2019)

\bibitem{newen2022unsupervised}
Newen, C., M{\"u}ller, E.: Unsupervised deepview: Global explainability of uncertainties for high dimensional data. In: 2022 IEEE International Conference on Knowledge Graph (ICKG). pp. 196--202. IEEE (2022)

\bibitem{pohler2016itemanalyse}
P{\"o}hler, G., Heine, T., Deml, B.: Itemanalyse und faktorstruktur eines fragebogens zur messung von vertrauen im umgang mit automatischen systemen. Zeitschrift f{\"u}r Arbeitswissenschaft  \textbf{3}(70),  151--160 (2016)

\bibitem{ribeiro2016should}
Ribeiro, M.T., Singh, S., Guestrin, C.: " why should i trust you?" explaining the predictions of any classifier. In: Proceedings of the 22nd ACM SIGKDD international conference on knowledge discovery and data mining. pp. 1135--1144 (2016)

\bibitem{ribeiro2018anchors}
Ribeiro, M.T., Singh, S., Guestrin, C.: Anchors: High-precision model-agnostic explanations. In: Proceedings of the AAAI conference on artificial intelligence. vol.~32 (2018)

\bibitem{ribera2019can}
Ribera, M., Lapedriza~Garc{\'\i}a, {\`A}.: Can we do better explanations? a proposal of user-centered explainable ai. CEUR Workshop Proceedings (2019)

\bibitem{rosenfeld2019explainability}
Rosenfeld, A., Richardson, A.: Explainability in human--agent systems. Autonomous Agents and Multi-Agent Systems  \textbf{33},  673--705 (2019)

\bibitem{salem2015would}
Salem, M., Lakatos, G., Amirabdollahian, F., Dautenhahn, K.: Would you trust a (faulty) robot? effects of error, task type and personality on human-robot cooperation and trust. In: Proceedings of the tenth annual ACM/IEEE international conference on human-robot interaction. pp. 141--148 (2015)

\bibitem{schulz2019deepview}
Schulz, A., Hinder, F., Hammer, B.: Deepview: Visualizing classification boundaries of deep neural networks as scatter plots using discriminative dimensionality reduction. arXiv preprint arXiv:1909.09154  (2019)

\bibitem{setzu2021glocalx}
Setzu, M., Guidotti, R., Monreale, A., Turini, F., Pedreschi, D., Giannotti, F.: Glocalx-from local to global explanations of black box ai models. Artificial Intelligence  \textbf{294},  103457 (2021)

\bibitem{seuss2021bridging}
Seu{\ss}, D.: Bridging the gap between explainable ai and uncertainty quantification to enhance trustability. arXiv preprint arXiv:2105.11828  (2021)

\bibitem{shin2021effects}
Shin, D.: The effects of explainability and causability on perception, trust, and acceptance: Implications for explainable ai. International Journal of Human-Computer Studies  \textbf{146},  102551 (2021)

\bibitem{siau2018building}
Siau, K., Wang, W.: Building trust in artificial intelligence, machine learning, and robotics. Cutter business technology journal  \textbf{31}(2),  47--53 (2018)

\bibitem{sjoberg1995overtraining}
Sj{\"o}berg, J., Ljung, L.: Overtraining, regularization and searching for a minimum, with application to neural networks. International Journal of Control  \textbf{62}(6),  1391--1407 (1995)

\bibitem{slack2021reliable}
Slack, D., Hilgard, A., Singh, S., Lakkaraju, H.: Reliable post hoc explanations: Modeling uncertainty in explainability. Advances in neural information processing systems  \textbf{34},  9391--9404 (2021)

\bibitem{sundararajan2017axiomatic}
Sundararajan, M., Taly, A., Yan, Q.: Axiomatic attribution for deep networks (2017)

\bibitem{tenhundfeld2019calibrating}
Tenhundfeld, N.L., De~Visser, E.J., Haring, K.S., Ries, A.J., Finomore, V.S., Tossell, C.C.: Calibrating trust in automation through familiarity with the autoparking feature of a tesla model x. Journal of cognitive engineering and decision making  \textbf{13}(4),  279--294 (2019)

\bibitem{toreini2020relationship}
Toreini, E., Aitken, M., Coopamootoo, K., Elliott, K., Zelaya, C.G., Van~Moorsel, A.: The relationship between trust in ai and trustworthy machine learning technologies. In: Proceedings of the 2020 conference on fairness, accountability, and transparency. pp. 272--283 (2020)

\end{thebibliography}

\end{document}